\documentclass[conference, letterpaper]{IEEEtran}
\IEEEoverridecommandlockouts
\usepackage{cite}
\usepackage{amsmath,amssymb,amsfonts}
\usepackage{graphicx}
\usepackage{textcomp}
\usepackage{xcolor}
\usepackage{graphics} 
\usepackage{epsfig} 
\usepackage{mathptmx} 
\usepackage{times} 
\usepackage{amsmath} 
\usepackage{amssymb}  
\usepackage{multirow}
\usepackage{array}
\usepackage{booktabs}
\usepackage{booktabs}
\usepackage{amsmath}
\usepackage[ruled,linesnumbered]{algorithm2e}
\usepackage{algpseudocode}
\usepackage{pgfplots}
\usepackage{mathptmx} 
\usepackage{amsmath}
\usepackage{comment}
\usepackage{graphicx}
\usepackage{booktabs}
\usepackage{bm}
\usetikzlibrary{pgfplots.groupplots}
\usetikzlibrary{patterns} 
\def\BibTeX{{\rm B\kern-.05em{\sc i\kern-.025em b}\kern-.08em T\kern-.1667em\lower.7ex\hbox{E}\kern-.125emX}}
\pgfplotsset{compat=1.18}

\begin{document}

\title{DP-LET : An Efficient Spatio-Temporal Network Traffic Prediction Framework\\
}

\author{
    \IEEEauthorblockN{
        Xintong Wang\IEEEauthorrefmark{1}, 
        Haihan Nan\IEEEauthorrefmark{2}, 
        Ruidong Li\IEEEauthorrefmark{1} and
        Huaming Wu\IEEEauthorrefmark{3}
    }
    \IEEEauthorblockA{
        \IEEEauthorrefmark{1}Kanazawa University, Kanazawa, Japan\\
        \IEEEauthorrefmark{2}George Mason University, Virginia, US\\
        \IEEEauthorrefmark{3}Tianjin University, Tianjin, China\\
        Email: 
            \IEEEauthorrefmark{1}wangxt@stu.kanazawa-u.ac.jp,
            \IEEEauthorrefmark{2}hnan@gmu.edu, 
            \IEEEauthorrefmark{1}liruidong@ieee.org,
            \IEEEauthorrefmark{3}whming@tju.edu.
    }
}

\maketitle
\begin{abstract}

Accurately predicting spatio-temporal network traffic is essential for dynamically managing computing resources in modern communication systems and minimizing energy consumption. Although spatio-temporal traffic prediction has received extensive research attention, further improvements in prediction accuracy and computational efficiency remain necessary. In particular, existing decomposition-based methods or hybrid architectures often incur heavy overhead when capturing local and global feature correlations, necessitating novel approaches that optimize accuracy and complexity. In this paper, we propose an efficient spatio-temporal network traffic prediction framework, DP-LET, which consists of a data processing module, a local feature enhancement module, and a Transformer-based prediction module. The data processing module is designed for high-efficiency denoising of network data and spatial decoupling. In contrast, the local feature enhancement module leverages multiple Temporal Convolutional Networks (TCNs) to capture fine-grained local features. Meanwhile, the prediction module utilizes a Transformer encoder to model long-term dependencies and assess feature relevance. A case study on real-world cellular traffic prediction demonstrates the practicality of DP-LET, which maintains low computational complexity while achieving state-of-the-art performance, significantly reducing MSE by 31.8\% and MAE by 23.1\% compared to baseline models.

\end{abstract}

\begin{IEEEkeywords}
Traffic Prediction, Deep Learning, Transformer, Performance Evaluation.
\end{IEEEkeywords}

\section{Introduction}
\label{sec:introduction}

The rapid proliferation of smart devices has led to the widespread adoption of the Internet and an explosion in data traffic and the number of applications. This exponential growth significantly enlarges the volume of data that must be collected and managed, increasing network complexity. In the Internet of Things (IoT) era and advanced wireless communications, these trends have further complicated network environments, making efficient management and optimization critical\cite{background1}.

Network traffic prediction is a vital technology designed to forecast the overall traffic volume based on historical data. By accurately predicting future traffic, network operators can preemptively mitigate network congestion and ensure high-quality service, which is crucial for devising effective resource allocation strategies and maintaining robust network performance. Traditionally, statistical methods—such as the Auto-Regressive Integrated Moving Average (ARIMA) model—have been used for traffic prediction. However, while these linear models can perform reasonably well in short-term scenarios, they often struggle with the complex and dynamic traffic patterns observed in large-scale IoT networks and modern communication systems\cite{background2}. As a result, deep learning techniques have emerged as a promising alternative to capture better the inherent non-linearities and long-range dependencies present in today's network traffic.

In deep learning, some studies have treated network traffic as images and leveraged Convolutional Neural Networks (CNNs) to capture and predict spatio-temporal dependencies \cite{cnn}. In contrast, others have employed Graph Neural Networks (GNNs) because they can learn topological information \cite{gnn}. However, CNNs rely on fixed receptive fields, making it challenging to model long-range temporal dependencies. Meanwhile, GNNs depend on predefined adjacency matrices, which poses challenges when accommodating dynamic node additions or removals \cite{edtp}. By contrast, Transformer-based methods can address these issues effectively, offering a more flexible approach to spatio-temporal traffic prediction through self-attention mechanisms that excel at handling long-range dependencies and adapting to evolving network topologies. Nevertheless, existing Transformer architectures suffer from two critical deficiencies that hinder them from achieving higher accuracy. Specifically, they exhibit insufficient sensitivity to local features—such as burst traffic patterns and periodic fluctuations in the temporal domain. Furthermore, they lack explicit spatial modeling mechanisms, leading to spatial confusion when predicting large-scale network topologies.

To further address Transformers' insufficient sensitivity to local features, current research primarily focuses on improving temporal feature extraction in the following two ways. The first approach decomposes raw data into multi-scale, seasonally informed components for separate predictions \cite{autoformer}, which inevitably increases computational overhead. The second approach adopts hybrid models that integrate additional networks with Transformers \cite{tcn-transformer}, but still faces an inherent trade-off between effective local feature capture and computational efficiency. Meanwhile, although techniques such as channel-independence \cite{patchtst} and modified attention mechanisms \cite{crossformer} have been proposed to handle spatial dependencies, they cannot bring significant breakthroughs in enhancing temporal feature extraction.

To address these challenges, we propose \textbf{DP-LET}, a spatio-temporal traffic prediction framework composed of a \textbf{D}ata \textbf{P}rocessing module, a \textbf{L}ocal \textbf{E}nhancement module, and a \textbf{T}ransformer-based prediction module. Specifically, the data processing module leverages Truncated Singular Value Decomposition and Reconstruction (TSVDR) combined with a spatial decoupling strategy, circumventing complex spatial modeling while reinforcing critical temporal features. Furthermore, the local feature enhancement module projects data into high-dimensional space. It employs Temporal Convolutional Networks (TCNs) with a dense layer to effectively capture multi-scale local patterns, compensating for the Transformer's limitations in local temporal dependency modeling. Finally, we employ Transformer as the backbone of the prediction module to capture global dependencies. DP-LET enhances prediction accuracy while maintaining computational efficiency. The main contribution of this study can be summarized as follows:

\begin{itemize}
\item We design a data processing module based on TSVDR with a spatial decoupling strategy. Specifically, TSVDR enables noise removal while preserving key temporal correlations, and explicit spatial decoupling eliminates the need for complex spatial modeling. Compared with the seasonal decomposition-based method, this module reduces GPU memory consumption by 38.4\%.

\item We develop a local feature enhancement module that maps network data into high-dimensional space and employs TCNs with dense layers to capture multi-scale local patterns. 

\item We integrate the data processing module, local feature enhancement module, and prediction module into the DP-LET framework. This architecture explicitly addresses spatial dependencies while modeling local and global interactions to capture spatio-temporal correlations. Evaluations of real-world cellular network traffic data demonstrate that DP-LET achieves significant performance gains, reducing the average MSE by 31.8\% and the average MAE by 23.1\% compared to state-of-the-art baselines.

\end{itemize}

The remainder of this paper is organized as follows. Section II introduces the recent literature on network traffic prediction. Section III presents the structure of DP-LET and describes each module in detail. In Section IV, a case study is presented. Section V concludes the paper by summarizing the main findings.

\section{Literature Review}

Recent research has addressed two critical challenges in Transformer-based spatio-temporal traffic prediction: local feature extraction, for example, detecting traffic bursts, and spatial dependency modeling, such as establishing node relationships.

To enhance temporal feature capture, Autoformer\cite{autoformer} pioneered seasonal decomposition to explicitly separate data into seasonal and trend components for prediction. This approach has been widely adopted by subsequent works\cite{edtp,fedformer}. However, since the seasonal and trend components must be predicted separately, the computational cost increases substantially. Some studies have proposed using simpler predictors —such as a linear layer\cite{dlinear} or multilayer perceptrons (MLPs)\cite{timemixer}—as a replacement for the Transformer to address the resulting parameter explosion. Although these approaches reduce some complexity, they do not fundamentally resolve the parameter explosion caused by decomposition.

A promising direction emerges from natural language processing (NLP), where Singularformer\cite{tsvdr} leverages TSVDR to reconstruct attention mechanisms, achieving significant complexity reduction. However, directly applying TSVDR to network traffic prediction faces a unique challenge. While the denoised results are remarkably effective in the temporal dimension, they disrupt the interrelationships among different spatial dimensions, making subsequent spatial modeling more challenging.

Model enhancement approaches have also emerged to mitigate the computational overhead introduced by decomposition-based strategies. Some researchers have explored hybrid architectures to leverage multiple neural network techniques within a single framework. For example, the CNN-LSTM-Transformer structure proposed in \cite{cnn-lstm-trans} uses sliding convolutional layers to reinforce local feature extraction. However, the depth of this network inevitably adds substantial computational overhead. One recent variant further replaces the CNN component with a TCN to refine local feature capture \cite{tcn-transformer}. However, these approaches still overlook explicit spatial relationships during spatio-temporal prediction.

For spatial dependency challenges, PatchTST\cite{patchtst} attempts spatial decoupling by treating variables as independent time series, while Crossformer\cite{crossformer} designs attention mechanisms to capture cross-variable dependencies explicitly. Nevertheless, these methods cannot address attention mechanisms' inherent temporal feature extraction limitations.

While existing methods either focus on improving temporal feature extraction or spatial dependency modeling, they fail to address both challenges simultaneously. Therefore, we propose DP-LET, an efficient spatio-temporal traffic prediction framework that simultaneously addresses the traditional Transformers' feature extraction deficiencies and spatial relationship modeling challenges while maintaining computational efficiency.

\begin{figure*}[h!]
    \centering
    \includegraphics[width=1\textwidth]{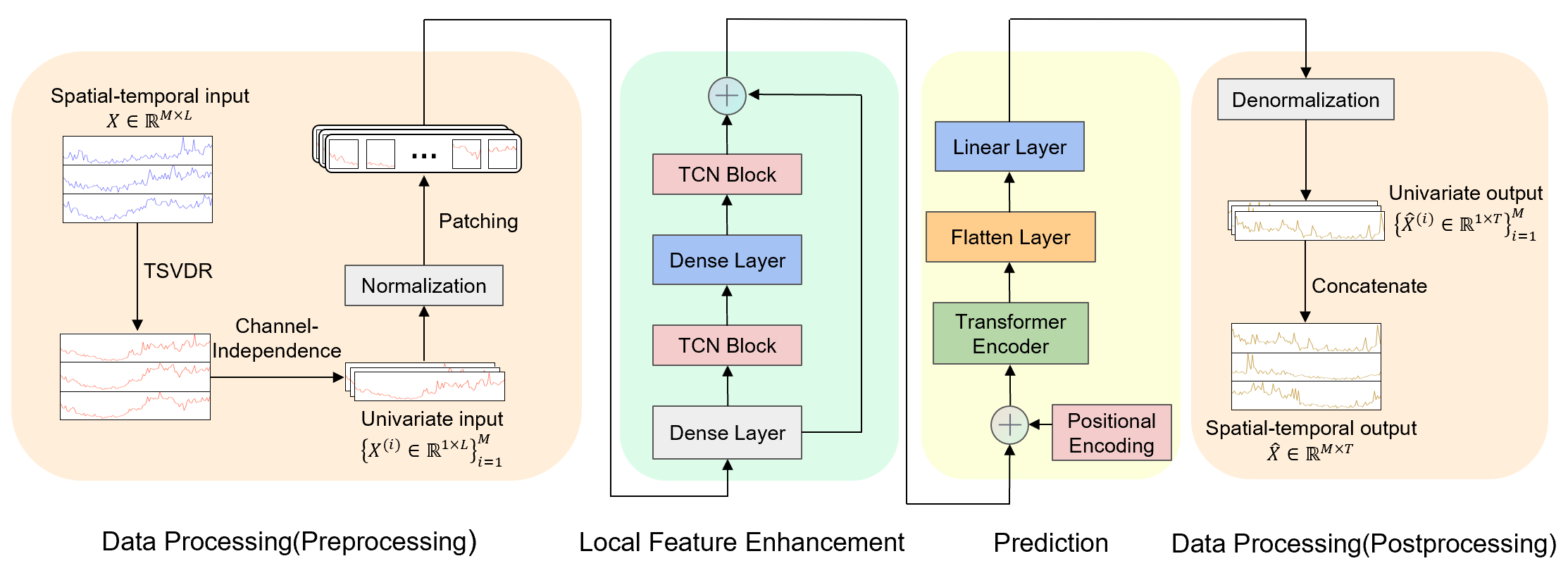}  
    \caption{DP-LET framework architecture}  
    \label{fig:framework}  
\end{figure*}

\section{DP-LET: An Efficient Framework for Spatio-Temporal Network Traffic Prediction}

DP-LET consists of three main modules: the data processing module, the local feature enhancement module, and the prediction module. The data processing module is subdivided into preprocessing and postprocessing parts. As shown in Fig.~\ref{fig:framework}, in the preprocessing stage, the raw data first undergoes TSVDR-based denoising and is then decoupled into multiple univariate time series. Afterward, these time series are normalized and segmented. Next, the local feature enhancement module refines local temporal dependencies within the resulting segments. The prediction module then employs a Transformer encoder to capture global dependencies and generate channel-wise forecasts. Finally, the postprocessing part restores the forecasts to their original scale and concatenates them into the final spatio-temporal predictions. In the following subsections, we describe each module in detail.

\subsection{Data Processing Module (Preprocessing Part)}
\label{subsec:preprocessing}

In the preprocessing stage, the input spatio-temporal data \({X} \in \mathbb{R}^{M \times L}\)is first denoised via TSVDR, where \(M\) denotes the number of channels and \(L\) denotes the input time horizon. Channel independence is applied to decouple the spatio-temporal data into \(M\) univariate time series \(\{X^{(i)} \in \mathbb{R}^{1 \times L}\}_{i=1}^{M}\) to address challenges in spatial relationship modeling that arise after TSVDR denoising. This approach circumvents the pitfalls of imposing artificial spatial relationships while preserving critical temporal dynamics. Reversible instance normalization and temporal patching mechanisms are incorporated to enhance data stationarity and capture multi-scale patterns, as illustrated in Algorithm\,\ref{alg:preprocessing}.

\begin{algorithm}[htb]
\caption{Data Processing Module (Preprocessing Part)}
\label{alg:preprocessing}
\KwData{\(X \in \mathbb{R}^{M \times L}\), threshold \(c \in \mathbb{R}_{+}\), segment length \(\ell\), stride \(S\)}
\KwResult{$\left\{ X_p^{(i)}[n] \in \mathbb{R}^{1 \times \ell}\right\}_{\substack{n=0,\dots,N-1 \\ i=1,\dots,M}}$, \(\{\mu_i,\sigma_i\}_{i=1}^{M}\)}

\(U\, \Sigma\, V^\top \gets \text{SVD}(X)\);

\For{\(j \gets 1\) \textbf{to} \(\min(M,L)\)}{
    \If{\(\Sigma_{j,j} < c\)}{
        \(\Sigma_{j,j} \gets 0\);
    }
}
\(X' \gets U\, \Sigma\, V^\top\);

\For{\(i \gets 1\) \textbf{to} \(M\)}{
    \(X^{(i)} \gets X'[i,:]\);
    
    \(\mu_i \gets \frac{1}{L}\sum_{t=1}^{L} X^{(i)}(t)\);
    
    \(\sigma_i \gets \sqrt{\frac{1}{L}\sum_{t=1}^{L}\bigl(X^{(i)}(t)-\mu_i\bigr)^2}\);

    \(X^{(i)}_{\text{norm}} \gets \frac{X^{(i)} - \mu_i}{\sigma_i}\);
    
    \(X^{(i)}_{\text{pad}} \gets \bigl[X^{(i)}_{\text{norm}},\, 0,\, 0,\, \dots,\, 0 \bigr]\);
    
    \(N \gets \left\lfloor \frac{L - \ell}{S} \right\rfloor + 2\);
    
    \For{\(n \gets 0\) \textbf{to} \(N-1\)}{
         \(X_p^{(i)}[n+1] \gets X^{(i)}_{\text{pad}}\bigl[n \cdot S + 1 : n \cdot S + \ell\bigr]\);
    }
}

\Return $\left\{ X_p^{(i)}[n] \} \right\}_{\substack{n=0,\dots,N-1 \\ i=1,\dots,M}},\, \{\mu_i,\sigma_i\}_{i=1}^{M}$;

\end{algorithm}

After the preprocessing stage, a total of \(N\) temporal segments of length \(\ell\) are obtained for each univariate channel. These segments can be expressed as $\left\{ X_p^{(i)}[n] \in \mathbb{R}^{1 \times \ell}\right\}_{\substack{n=0,\dots,N-1 \\ i=1,\dots,M}}$, where each segment captures local temporal dynamics within a fixed window of length \(\ell\). The resulting representation $\left\{ X_p^{(i)}[n] \}\right\}_{\substack{n=0,\dots,N-1 \\ i=1,\dots,M}}$ not only encompasses the denoised, normalized, and segmented data but also preserves the essential characteristics of the original time series, thereby laying a solid foundation for efficient multi-scale feature extraction and forecasting in subsequent stages.

\subsection{Local Feature Enhancement Module}

Multi-branch TCNs serve as the backbone of this module. Compared to CNNs, TCNs can effectively capture local features via dilated convolutions, reducing the required network depth. Initially, a dense layer projects each segment $\left\{ X_p^{(i)}[n] \}\right\}_{\substack{n=0,\dots,N-1 \\ i=1,\dots,M}}$ to a new representation $\left\{\tilde{X}_p^{(i)}[n] \in \mathbb{R}^{1 \times d_{\text{model}}}\right\}_{\substack{n=0,\dots,N-1 \\ i=1,\dots,M}}$, thus enhancing the representational capacity of the input.The projected segment $\left\{\tilde{X}_p^{(i)}[n] \}\right\}_{\substack{n=0,\dots,N-1 \\ i=1,\dots,M}}$ is then processed by a TCN block to strengthen local feature extraction further, followed by a dense layer that refines and recalibrates the learned features. Subsequently, a second TCN block is applied to capture additional temporal dynamics. Finally, a residual connection is introduced to mitigate issues such as gradient explosion and to ensure stable training. The final output of this module is $\left\{X_e^{(i)}[n] \in \mathbb{R}^{1 \times d_{\text{model}}}\right\}_{\substack{n=0,\dots,N-1 \\ i=1,\dots,M}}$

The framework incorporates this module as an integral component, enabling it to seamlessly replace the conventional Transformer data embedding layer and be readily generalized to other Transformer-based architectures.

\subsection{Prediction Module}

A classical Transformer encoder is employed in the prediction module to model long-range temporal dependencies and generate final forecasts. A learnable positional encoding is then added to $\left\{X_e^{(i)}[n] \}\right\}_{\substack{n=0,\dots,N-1 \\ i=1,\dots,M}}$, and the sequence is fed into the Transformer encoder, retaining the same shape while capturing global temporal dependencies across all segments. Next, the encoder output is flattened into \(\{X_{\text{flat}}^{(i)} \in \mathbb{R}^{1 \times (N \cdot d_{\text{model}})}\}_{i=1}^{M}\), thereby consolidating information from all segments. Finally, a linear layer projects \(\{X_{\text{flat}}^{(i)}\}_{i=1}^{M}\) to \(\{{X}_{\text{linear}}^{(i)} \in \mathbb{R}^{1 \times T}\}_{i=1}^{M}\), where \(T\) is the prediction horizon.

\subsection{Data Processing Module (Postprocessing Part)}
Since the preprocessing transformations alter the original data structure, corresponding inverse operations must project the model predictions back to the original data space. In the postprocessing phase, a denormalization step is applied to restore the predictions to their original scale. Subsequently, the univariate prediction results for all channels, \(\{\hat{X}^{(i)} \in \mathbb{R}^{1 \times T}\}_{i=1}^{M}\), are concatenated to reconstruct the final spatio-temporal prediction result \(\hat{X} \in \mathbb{R}^{M \times T}\). The detailed process is described in Algorithm~\ref{alg:postprocessing}.

\begin{algorithm}[htb]
\caption{Data Processing Module (Postprocessing Part)}
\label{alg:postprocessing}
\KwData{\(\{{X}_{\text{linear}}^{(i)} \in \mathbb{R}^{1 \times T}\}_{i=1}^{M}\), \(\{\mu_i,\sigma_i\}_{i=1}^{M}\)}
\KwResult{\(\hat{X} \in \mathbb{R}^{M \times T}\)}
\For{\(i \gets 1\) \textbf{to} \(M\)}{
    \(\hat{X}^{(i)} \gets {X}_{\text{linear}}^{(i)} \cdot \sigma_i + \mu_i\);
}
\(\hat{X} \gets \mathbf{0}^{M \times T}\);

\For{\(i \gets 1\) \textbf{to} \(M\)}{
    \(\hat{X}[i, :] \gets \hat{X}^{(i)}\);
}
\Return \(\hat{X}\);
\end{algorithm}

\section{A Case Study: Real-world Cellular Traffic Prediction}
Cellular network traffic is a critical data type underpinning modern communication systems, serving billions of mobile users and devices. The evolution of cellular networks—from the first-generation (1G) analog systems to the sophisticated 5G networks—has led to dramatic increases in data transmission rates and service complexity. In this context, accurate traffic prediction becomes essential for optimizing network performance, ensuring reliable service, and enabling strategic operations such as failure detection, adaptive base station sleep scheduling, and efficient resource allocation\cite{cellular}.

In this case study, we leverage a real-world cellular network traffic dataset to validate the effectiveness of our proposed framework. We begin by outlining the fundamental characteristics of cellular traffic data. Next, we compare DP-LET with other state-of-the-art methods regarding prediction accuracy and computational complexity. To further assess the contribution of each module, we conduct a series of ablation experiments analyzing their impact on accuracy and complexity. Finally, we evaluate the transferability of the local enhancement module, demonstrating its potential for seamless integration into other Transformer-based architectures.

\subsection{Experimental Setting}

 \textit{1) Dataset:} In this study, we utilized the Call Detail Records (CDRs) dataset \cite{milan} collected by Telecom Italia in Milan, which represents the mobile network traffic data for the city of Milan. The data were acquired at a 10-minute interval, yielding 8,928 time stamps that chronicle the evolution of mobile network traffic from November 1, 2013, to January 1, 2014 (62 consecutive days). Spatially, Milan is divided into a 100×100 grid, resulting in 10,000 equally-sized regions. The spatio-temporal characteristics of the dataset are detailed in Table~\ref{tab:table1}.

\begin{table}[t!]
\caption{Spatio-temporal characteristics of the Milan CDRs dataset}
\label{tab:table1}
\centering
\begin{tabular}{@{}lccc@{}}
\toprule
Dimension & Data Size & Range & Minimum unit \\ 
\midrule
Spatial & \(100 \times 100\) & 10,000 grids & Per-grid network traffic \\
Temporal & 8,928 & 62 days & 10 minutes \\
\bottomrule
\end{tabular}
\end{table}

\addtolength{\topmargin}{+0.15cm}
\begin{table*}[t]
\caption{Performance comparison of DP-LET and baseline methods}
\label{tab:mainresult}
\centering
\scalebox{0.85}{
\begin{tabular}{l|c@{\hskip 3pt}c|c@{\hskip 3pt}c|c@{\hskip 3pt}c|c@{\hskip 3pt}c|c@{\hskip 3pt}c|c@{\hskip 3pt}c|c@{\hskip 3pt}c|c@{\hskip 3pt}c|c@{\hskip 3pt}c|c@{\hskip 3pt}c}
\toprule
Methods & \multicolumn{2}{c}{\textbf{DP-LET}} 
          & \multicolumn{2}{c}{TimeMixer}
          & \multicolumn{2}{c}{PatchTST}
          & \multicolumn{2}{c}{iTransformer}
          & \multicolumn{2}{c}{TimesNet}
          & \multicolumn{2}{c}{DLinear}
          & \multicolumn{2}{c}{FEDformer}
          & \multicolumn{2}{c}{Autoformer}
          & \multicolumn{2}{c}{Crossformer}
          & \multicolumn{2}{c}{Informer} \\
\cmidrule(lr){2-3} \cmidrule(lr){4-5} \cmidrule(lr){6-7} \cmidrule(lr){8-9}
\cmidrule(lr){10-11} \cmidrule(lr){12-13} \cmidrule(lr){14-15}
\cmidrule(lr){16-17} \cmidrule(lr){18-19} \cmidrule(lr){20-21}
Metrics & MSE & MAE
& MSE & MAE
& MSE & MAE
& MSE & MAE
& MSE & MAE
& MSE & MAE
& MSE & MAE
& MSE & MAE
& MSE & MAE
& MSE & MAE \\
\midrule
\(T=72\)
& \textbf{0.225} & \textbf{0.324} 
& \underline{0.230} & \underline{0.329} 
& 0.234 & 0.337
& 0.237 & 0.333
& 0.304 & 0.385
& 0.311 & 0.428
& 0.387 & 0.468
& 0.429 & 0.506
& 0.510 & 0.517
& 0.653 & 0.615 \\
\(T=144\)
& \textbf{0.238} & \textbf{0.333}
& 0.255 & 0.347
& \underline{0.249} & 0.350
& 0.256 & \underline{0.346}
& 0.341 & 0.410
& 0.335 & 0.447
& 0.408 & 0.485
& 0.443 & 0.513
& 0.646 & 0.611
& 0.749 & 0.654 \\
\bottomrule
\end{tabular}}
\end{table*}

\pgfplotsset{compat=1.18}

\textit{2) Evaluation criteria:} In prediction tasks, Mean Squared Error (MSE) and Mean Absolute Error (MAE) are commonly used metrics because they assess regression performance from different perspectives. These metrics are adopted in this paper to evaluate prediction accuracy comprehensively. Their formulas are defined as follows:

\begin{equation}
\text{MSE} = \frac{1}{N} \sum_{i=1}^{N} (y_i - \hat{y}_i)^2,
\end{equation}
\begin{equation}
\text{MAE} = \frac{1}{N} \sum_{i=1}^{N} \left| y_i - \hat{y}_i \right|,
\end{equation}

where \(N\) denotes the total number of samples, \(y_i\) represents the actual network traffic, and \(\hat{y}_i\) represents the predicted network traffic. In addition, the number of parameters is reported to provide insight into the computational complexity of the proposed framework.

\textit{3) Baselines:} We compare our framework with nine state-of-the-art models, including six leading Transformer-based methods: iTransformer \cite{itransformer}, Informer \cite{informer}, PatchTST, Autoformer, FEDformer and Crossformer; in addition, we consider the CNN-based model TimesNet \cite{timesnet}, the linear model DLinear, and the MLP-based model TimeMixer. All of these models have demonstrated strong performance in spatio-temporal prediction tasks.

\textit{4) Implementation details:} All experiments were conducted on an NVIDIA RTX 5000 GPU. We randomly selected 100 spatial grids from Milan's \(100 \times 100\) cellular network to ensure a representative subset of the overall traffic distribution. The temporal configuration adopts a 10-minute resolution, with input sequences spanning three consecutive days (\(L = 432\) time steps) to predict two distinct horizons: 12-hour (\(T = 72\)) and 24-hour (\(T = 144\)) network traffic flows. 

All baseline models were implemented using their officially reported hyperparameters. The maximum number of training epochs was set to 100, with an early stopping patience of 20 epochs.

\subsection{Experimental Result}

\textit{1) Main results:} We conduct experiments on a cellular network traffic dataset and compare our proposed framework with multiple baseline methods under two prediction horizons, \(T \in \{72, 144\}\). As shown in Table~\ref{tab:mainresult}, the best performance is highlighted in \textbf{bold}, and the second-best result is \underline{underlined}. DP-LET consistently outperforms the baselines at both \(T=72\) and \(T=144\). On average, it achieves a 31.8\% reduction in MSE and a 23.1\% reduction in MAE compared to the other methods, underscoring its superior predictive capabilities.

\begin{figure}[t!]
  \centering
  \resizebox{1\linewidth}{!}{%
  \begin{tikzpicture}

    \definecolor{ieeeblue}{RGB}{0,98,155}   
    \definecolor{ieeegold}{RGB}{198,146,20} 

\tikzset{
  barA/.style={
    ybar, draw=black, line width=0.4pt, fill=ieeeblue,
    postaction={pattern=north east lines, pattern color=black}
  },
  barB/.style={
    ybar, draw=black, line width=0.4pt, fill=ieeegold,
    postaction={pattern=north west lines, pattern color=black}
  }
}

    \begin{axis}[
      name=ax1,
      ybar,
      bar width=12pt,
      bar shift=-6pt,
      x=1cm,
      ymode=log, log basis y=10,
      ymin=5e4,
      ylabel={\small Number of Parameters},
      ytick={1e5,1e6,1e7,1e8},
      yticklabels={$10^5$,$10^6$,$10^7$,$10^8$},
      symbolic x coords={
        DLinear,iTransformer,PatchTST,TimeMixer,DP-LET,
        Autoformer,Informer,Fedformer,TimesNet,Crossformer
      },
      xtick=data,
      xticklabel style={font=\small,rotate=45,anchor=east},
      enlarge x limits=0.1,
      legend style={
        font=\small,
        at={(0.03,0.97)},anchor=north west,
        cells={anchor=west}
      }
    ]
      \addplot[barA]
        coordinates {
          (DLinear,62352) (iTransformer,264136) (PatchTST,897480)
          (TimeMixer,1732673) (DP-LET,1847128) (Autoformer,13342180)
          (Informer,16319332) (Fedformer,21961572)
          (TimesNet,37761756) (Crossformer,44324388)
        };
      \addlegendentry{Number of Parameters}

      \addlegendimage{ybar,barB}
      \addlegendentry{Time per Training Iteration}
    \end{axis}
    
    \begin{axis}[
      at={(ax1.south west)}, anchor=south west,
      ybar,
      bar width=12pt,
      bar shift=+6pt,
      x=1cm,
      ymin=0, ymax=7,
      axis y line*=right,
      axis x line=none,
      ylabel={\small Time per Training Iteration (s)},
      symbolic x coords={
        DLinear,iTransformer,PatchTST,TimeMixer,DP-LET,
        Autoformer,Informer,Fedformer,TimesNet,Crossformer
      },
      xtick=data,
      xticklabels={},
      enlarge x limits=0.1
    ]
      \addplot[barB]
        coordinates {
          (DLinear,0.0401) (iTransformer,0.0586) (PatchTST,0.32)
          (TimeMixer,0.1597) (DP-LET,0.42) (Autoformer,0.3381)
          (Informer,0.697)   (Fedformer,0.8385)
          (TimesNet,6.39)    (Crossformer,2.2)
        };
    \end{axis}
  \end{tikzpicture}%
  }
  \caption{Efficiency analysis of DP-LET and baseline methods}
  \label{fig:para}
\end{figure}

Beyond predictive performance, we also evaluate the model complexity of each method by comparing the number of trainable parameters and time per training iteration, as shown in Fig.~\ref{fig:para}. Among Transformer-based approaches, DP-LET has the third-smallest parameter count. It substantially reduces both parameters and iteration time relative to seasonal-decomposition Transformers such as Autoformer and FEDformer, and its complexity is close to that of the MLP-based TimeMixer. These results demonstrate that our approach achieves high forecasting accuracy while maintaining a relatively light model footprint, effectively balancing efficiency and performance.

\textit{2) Ablation study:} We perform an ablation study to evaluate the contributions of the local feature enhancement module and the data processing module in our proposed framework for a prediction horizon of \(T = 144\). Specifically, we compare three configurations: (i) the prediction module combined with the data processing module, (ii) the prediction module combined with the local feature enhancement module, and (iii) the complete framework that integrates all modules. As shown in Table~\ref{tab:ablation}, the complete framework achieves the highest accuracy. Furthermore, compared with the local feature enhancement module, the data processing module contributes more substantially to prediction accuracy, as reflected by the lower error metrics when integrated with the prediction module.

\begin{table}[b!]
\caption{Results of ablation experiments}
\label{tab:ablation}
\centering
\begin{tabular}{lcc}
\toprule
Methods & MSE & MAE \\
\midrule
Prediction+Data Processing       & 0.254      & 0.349      \\
Prediction+Local Feature Enhancement      & 0.343      & 0.423      \\
\textbf{Proposed Framework} & \textbf{0.238}  & \textbf{0.333}  \\
\bottomrule
\end{tabular}
\end{table}

\begin{table}[b!]
\caption{Effect of the data processing module}
\label{tab:decomposition}
\centering
\scalebox{0.9}{
\begin{tabular}{lcccc}
\toprule
Metrics
& \multicolumn{2}{c}{Complexity}
& \multicolumn{2}{c}{Accuracy}  \\
\cmidrule(lr){2-3}\cmidrule(lr){4-5}
Methods & Para. & Mem. (MiB)
& MSE & MAE\\
\midrule
Seasonal Decomposition                 
& 3,694,456 & 13,485 & 0.263 &\textbf{0.349}\\
\midrule
\textbf{Proposed Data Processing}         
& \textbf{1,847,128} & \textbf{8,308} & \textbf{0.254} & \textbf{0.349}\\

\bottomrule
\end{tabular}
}
\end{table}

\begin{table*}[t!]
\caption{Effect of the local feature enhancement module}
\label{tab:local}
\centering
\begin{tabular}{lcccccccc}
\toprule
Methods
& \multicolumn{2}{c}{Transformer}
& \multicolumn{2}{c}{Non-stationary}
& \multicolumn{2}{c}{FEDformer}
& \multicolumn{2}{c}{Reformer} \\
\cmidrule(lr){2-3}\cmidrule(lr){4-5}\cmidrule(lr){6-7}\cmidrule(lr){8-9}
Metrics & MSE & MAE
& MSE & MAE
& MSE & MAE
& MSE & MAE \\
\midrule
Original                   
& 0.689 & 0.624 
& 0.414 & 0.457
& 0.408 & 0.485
& 0.632 & 0.594 \\
+ Local Feature Enhancement
& 0.606 & 0.588 
& 0.373 & 0.437
& 0.388 & 0.471
& 0.620 & 0.586 \\
\midrule
Improvement (\%)           
& 12.05\% & 5.77\%
& 9.90\% & 4.37\%
& 4.90\% & 2.89\%
& 1.90\% & 1.35\% \\
\bottomrule
\end{tabular}
\end{table*}

\textit{3) Effect of the data processing module:}  
To validate the effectiveness of our proposed data processing module, we integrated both the proposed module and a conventional seasonal decomposition approach into the prediction module and assessed their performance in terms of prediction accuracy and computational complexity. As shown in Table~\ref{tab:decomposition}, our method achieves a 3.42\% reduction in MSE relative to the baseline. Meanwhile, the proposed module requires only 50\% of the parameters (Para.) and 61.6\% of the GPU memory (Mem.) compared to seasonal decomposition, underscoring its superior efficiency.

\textit{4) Effect of the local feature enhancement module:}
We applied the proposed local feature enhancement module to four Transformer-based models at a prediction horizon of \(T=144\). 
Table~\ref{tab:local} reports the MSE and MAE values before and after integrating our module. 
Notably, each method sees a clear performance boost, with the Transformer experiencing the largest gains, 
followed by Nonstationary-Transformer\cite{nonstation}, Reformer\cite{reformer}, and FEDformer.
These findings confirm the effectiveness of capturing local dependencies for improved spatio-temporal forecasting.

\section{Conclusion}
This paper introduced DP-LET as a novel and efficient spatio-temporal network traffic prediction framework. The proposed approach integrates a robust data processing module for noise reduction and spatio-temporal data decoupling, a local feature enhancement module for capturing multi-scale temporal dependencies, and a Transformer-based prediction module for extracting global temporal patterns. This unified design strikes an effective balance between computational complexity and predictive accuracy.

Extensive experiments on a real-world cellular network traffic dataset demonstrate that DP-LET consistently outperforms state-of-the-art baselines across various forecasting horizons. Moreover, DP-LET's modular architecture not only boosts prediction performance but also underscores the exceptional portability of the local feature enhancement module, enabling seamless integration into other Transformer-based models and broadening its applicability to diverse prediction frameworks. 

\section*{Acknowledgment}

This work was partly supported by Japan Society for the Promotion of Science (JSPS) KAKENHI Grant Number 23K28070.

\bibliographystyle{IEEEtran}

\end{document}